\documentclass[letterpaper, 10 pt, conference]{ieeeconf}
\IEEEoverridecommandlockouts
\usepackage{graphicx}
\usepackage{subcaption}
\usepackage{float}
\usepackage{amsmath}
\usepackage{amssymb}
\usepackage{hyperref}
\hypersetup{hidelinks}
\usepackage{makecell}
\usepackage{booktabs}
\usepackage[table]{xcolor}
\usepackage{colortbl}
\usepackage{array}
\usepackage{multirow}
\usepackage{diagbox}
\usepackage{pgfplots}
\pgfplotsset{compat=1.18}
\usepgfplotslibrary{statistics}

\newif\iffastcompile
\fastcompilefalse
\iffastcompile
  \setkeys{Gin}{draft}
\fi

\definecolor{natureTeal}{RGB}{0, 128, 128}      % Inputs (Sensors)
\definecolor{natureGrey}{RGB}{112, 128, 144}    % Networks (Black box)
\definecolor{natureBlue}{RGB}{0, 51, 102}       % Logic (Algorithms)
\definecolor{natureSalmon}{RGB}{250, 128, 114}  % Execution (Action)
\definecolor{lightTeal}{RGB}{179, 224, 224}
\definecolor{lightGrey}{RGB}{193, 202, 213}
\definecolor{lightBlue}{RGB}{179, 198, 224}
\definecolor{lightSalmon}{RGB}{252, 196, 186}

\definecolor{SageGreenBG}{RGB}{238,244,236}
\definecolor{MistBlueBG}{RGB}{239,247,249}
\definecolor{ApricotBG}{RGB}{253,249,240}
\definecolor{SageGreenHL}{RGB}{217,229,212}
\definecolor{MistBlueHL}{RGB}{215,235,240}
\definecolor{ApricotHL}{RGB}{245,234,208}

\PassOptionsToPackage{shortlabels}{enumitem}
\let\labelindent\relax
\usepackage{enumitem}

\title{\LARGE \bf
R2HandoverSim: A Simulation Framework and Benchmark for Robot-to-Human Object Handovers
}

\author{Hanxin Zhang$^{1,2}$, Abdulqader Dhafer$^{1,2}$, Hongbiao Dong$^{3}$, Zhou Daniel Hao$^{1,2}$%
\thanks{$^{1}$Authors are with DANiLab, University of Leicester, Leicester, UK
{\tt\small \{hz273,aamd2,d.hao\}@leicester.ac.uk}.}%
\thanks{$^{2}$Authors are with the School of Computing and Mathematical Sciences, University of Leicester, Leicester, UK.}%
\thanks{$^{3}$The author is with the School of Metallurgy and Materials, University of Birmingham, Birmingham, UK.
{\tt\small h.dong.1@bham.ac.uk}.}%
}

\begin{document}

\makeatletter
\maketitle
\makeatother

\begin{abstract}
We present R2HandoverSim, a simulation benchmark for robot-to-human (R2H) object handovers.
Although R2H handover methods have advanced rapidly, the lack of standardized evaluation protocols impedes objective comparison.
Our benchmark enables reproducible evaluation by systematically comparing four baselines on their predicted shared grasp poses.
We conduct a user study with 30 participants, analyze baseline performance, and show that simulation results correlate with real-world evaluation outcomes.
Crucially, five complementary metrics (planning feasibility, reachability, grasp stability, grasp affordance, and safety) better reflect user-perceived handover quality than overall success rate alone.
Website and code: \url{https://robot-future.github.io/r2handoversim/}.
\end{abstract}

\section{Introduction}

Robot-to-human (R2H) object handover, in which a robot transfers an object to a human receiver, is a fundamental capability of service, assistive, and collaborative robotic systems~\cite{ortenziObjectHandoversReview2021}.
Recent methods focus on predicting a \emph{shared grasp pose}, the configuration in which the robot presents the object so that the human receiver can grasp it.
Approaches to this problem span heuristic optimization~\cite{lehotsky2023optimizing}, contact prediction~\cite{wang2024contacthandover}, and generative models~\cite{laplaza2021attention};
yet they differ substantially in input modalities, object sets, and evaluation protocols.
Because no standardized benchmark exists, comparisons across R2H methods ultimately rely on user studies whose results are sensitive to participant variability and difficult to reproduce.

In the human-to-robot (H2R) direction, simulation benchmarks such as HandoverSim~\cite{chao2022handoversim} and GenH2R~\cite{wang2024genh2r} have already established shared evaluation standards and accelerated progress.
In parallel, datasets that capture interactions between hands and objects~\cite{ye2021h2o, taheriGRABDatasetWholebody2020, fan2023arctic} provide reference hand configurations that can define handover goals.
Unlike H2R handovers, the R2H direction still lacks an analogous simulation benchmark for standardized evaluation.

Constructing such a benchmark is nontrivial because R2H success is harder to operationalize than H2R success.
In H2R handovers, success is typically measured by the robot's ability to grasp and retain an object, whereas R2H success depends on safe and comfortable reception by the human, whose adaptive behavior is difficult to model in simulation.
However, most R2H methods converge on predicting a shared grasp pose, making simulation evaluation feasible: rather than modeling full human grasping behavior, one can assess whether each predicted pose satisfies geometric, kinematic, and safety constraints in a physics simulator.

\begin{figure}[!t]
    \centering
    \includegraphics[width=\columnwidth]{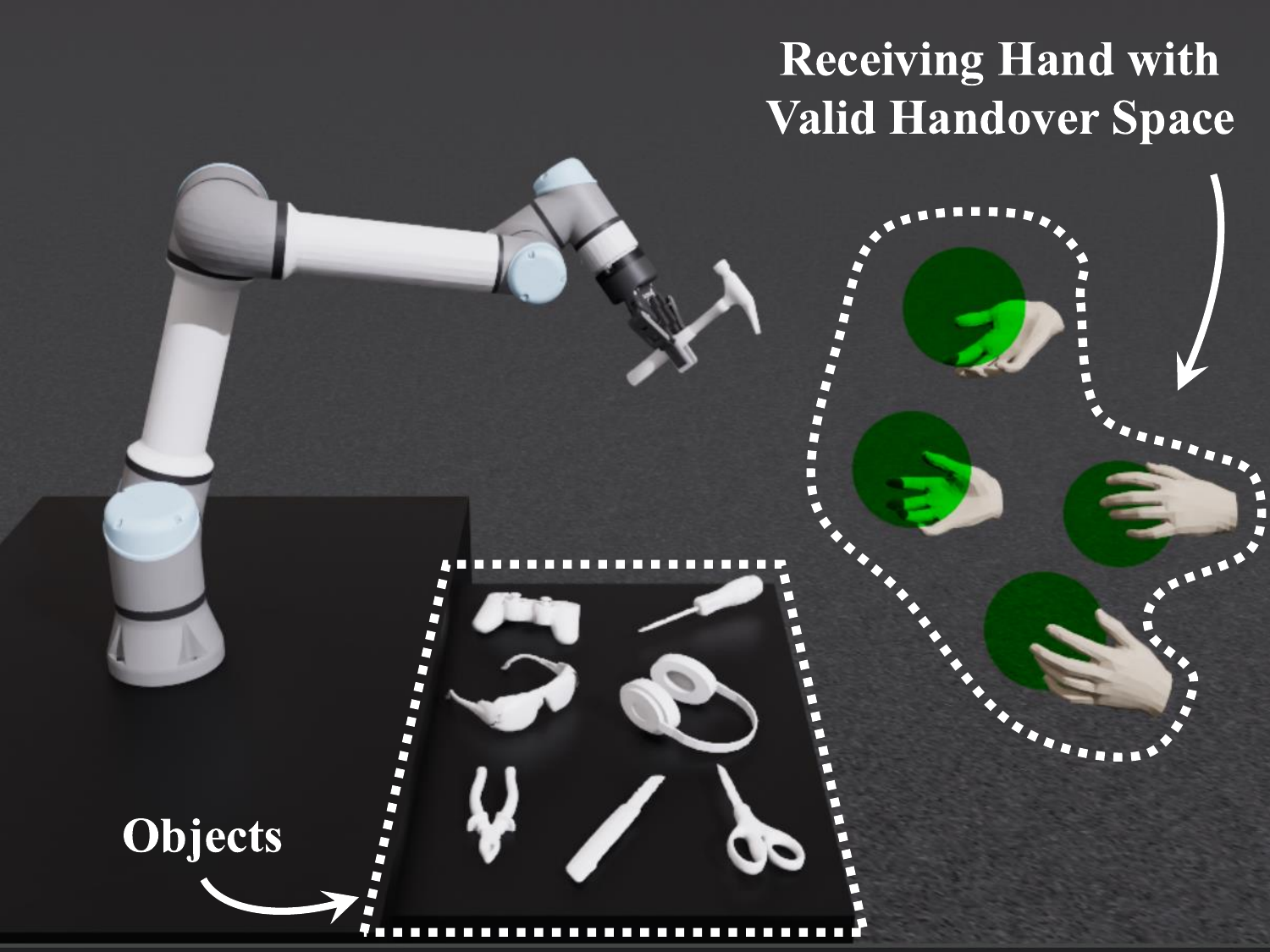}
    \caption{Overview of the R2HandoverSim benchmark environment. A UR5e manipulator delivers an object to a static MANO hand. The green sphere denotes the valid handover space used for reachability evaluation.}
    \label{fig:teaser}
    \vspace{-10pt}
\end{figure}

In this paper, we make three contributions.
\textbf{(1)}~We introduce R2HandoverSim (Fig.~\ref{fig:teaser}), a simulation benchmark that enables standardized and reproducible evaluation of R2H handover methods across diverse objects and evaluation settings.
\textbf{(2)}~We define an evaluation protocol with five complementary criteria (planning feasibility, reachability, grasp stability, grasp affordance, and safety) and use it to systematically compare four baselines.
\textbf{(3)}~We conduct a user study with 30 participants, showing that simulation evaluation identifies the top method in real world deployment while subjective ratings reveal additional cross method differences, validating the benchmark's practical relevance.

\begin{figure*}[!t]
\centering
\includegraphics[width=\textwidth]{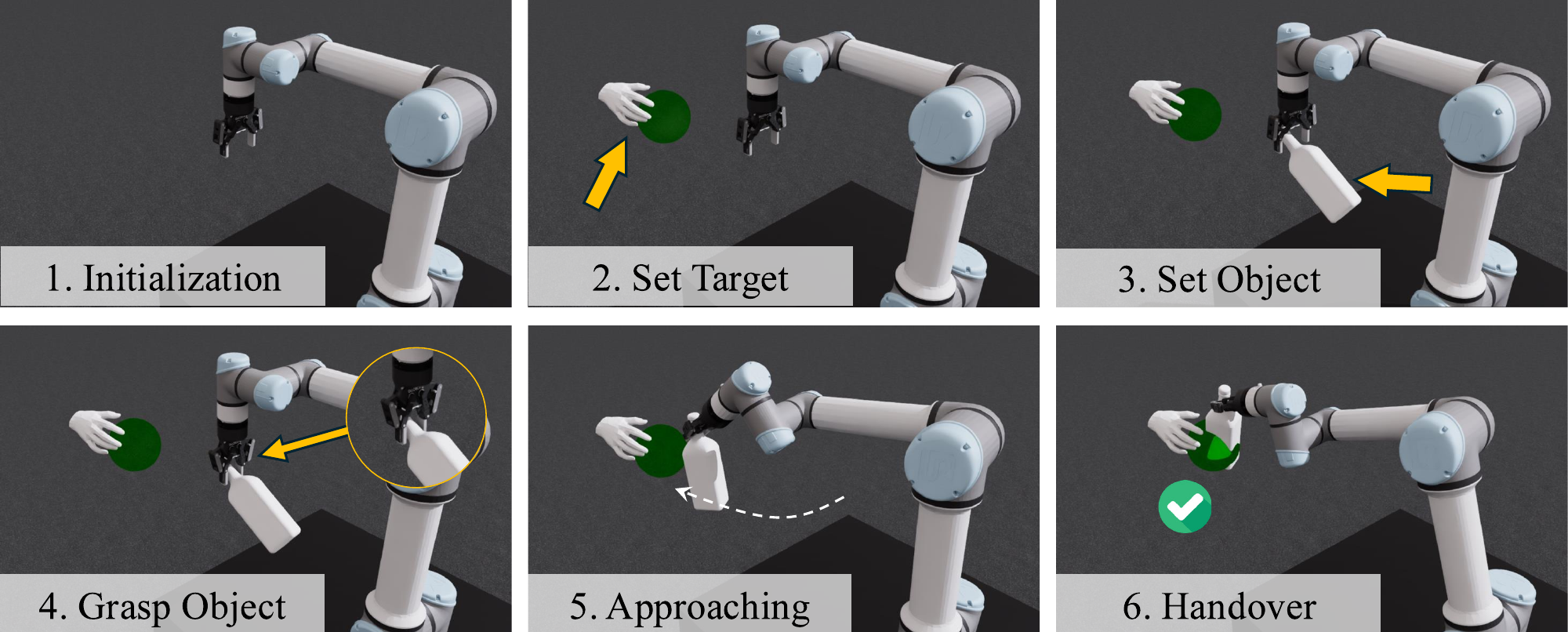}
\caption{Trial protocol of R2HandoverSim: initialization, target pose assignment, object placement, robotic grasping, approach motion, and handover evaluation.}
\label{fig:trial_protocol}
\vspace{-10pt}
\end{figure*}

\section{Related Work}

\noindent\textbf{Robot-to-Human Handovers.}
The goal of R2H handover is to present an object so that the receiver can grasp it safely and use it immediately \cite{duan2024human, ortenzi2020grasp}.
Existing R2H studies can be divided into methods for general handover and methods for task-specific handover.
General methods predict shared grasp poses through rule-based optimization \cite{lehotsky2023optimizing},
contact prediction \cite{wang2024contacthandover},
semantic and geometric reasoning \cite{liu2024leveraging},
and human pose estimation \cite{laplaza2021attention}.
Task-specific methods target scenarios such as assistive handover for older adults \cite{9802960} and bimanual object transfer~\cite{ovur2023naturalistic}.
However, these works vary widely in object sets (from a single item to over 30 household objects), evaluation protocols (simulation only, real robot trials, or user studies with different subjective scales), and metrics (from success rate to biomechanical comfort indices), making direct comparison difficult.
R2HandoverSim addresses this gap with a unified simulation environment built on a UR5e robot with a Robotiq 2F-85 gripper and 16 objects, together with five evaluation criteria covering planning feasibility, reachability, grasp stability, grasp affordance, and safety (Section~\ref{sec:metrics}), allowing R2H methods to be compared on equal footing for the first time.

\noindent\textbf{Handover Benchmark.}
In the H2R direction, benchmarks have already driven rapid progress.
Hand-object interaction datasets such as H2O \cite{ye2021h2o},
GRAB \cite{taheriGRABDatasetWholebody2020}, ARCTIC \cite{fan2023arctic}, ContactPose \cite{brahmbhatt2020contactpose}, and DexYCB \cite{chao2021dexycb} provide reference grasps for evaluating grasp feasibility,
while simulation benchmarks such as HandoverSim \cite{chao2022handoversim}, GenH2R \cite{wang2024genh2r},
MobileH2R \cite{wang2025mobileh2r}, and DexH2R \cite{wang2025dexh2r} offer standardized environments for H2R policies.
These resources share two ingredients transferable to R2H: curated hand-object interaction data defining plausible handover goals, and physics simulation enabling reproducible evaluation.
No benchmark yet combines them for R2H.
R2HandoverSim fills this gap by integrating hand-object interaction sequences from established datasets with robotic grasp candidates in a physics simulation, providing the first standardized R2H \emph{simulation} benchmark.

\section{R2HandoverSim: The Benchmark}

\subsection{Task Formulation}\label{sec:task}

We consider the robot-to-human (R2H) object handover task, in which a robotic manipulator transfers an object to a human receiver at a target handover pose.
At the start of each trial, the robot is initialized at a fixed home configuration $q_0 \in \mathbb{R}^n$, where $n$ denotes the number of actuated joints.
The receiver hand pose $T_{\text{hand}} \in SE(3)$ is randomly sampled per trial from a reachable set and then fixed in the world frame for that trial; its position $\mathbf{p}_h \in \mathbb{R}^3$ and outward palm normal $\mathbf{n}_h \in \mathbb{R}^3$ are extracted for computing evaluation criteria.

Each method outputs two end-effector poses: the grasp pose $T_{ee}^{g} \in SE(3)$ at which the robot holds the object, and the handover pose $T_{ee}^{h} \in SE(3)$ at which the object is delivered to the receiver.
The benchmark does not restrict how these poses are generated.
Once $T_{ee}^{g}$ is specified, the object is placed relative to the end-effector and the gripper closes.
If the object width along the closing axis exceeds the maximum aperture, the trial is recorded as a drop failure; otherwise, the object is rigidly attached for subsequent planning and execution.

Starting from $q_0$, the robot executes an inverse kinematics and motion planning pipeline to reach $T_{ee}^{h}$.
Inverse kinematics is solved with a numerical Jacobian iterative solver; motion planning uses RRT-Connect within MoveIt with collision checking enabled for the robot arm, end-effector, attached object, and MANO hand mesh.
The complete trial protocol is illustrated in Fig.~\ref{fig:trial_protocol}.
Each trial is evaluated based on the following five criteria.
\begin{itemize}
\item \textbf{Plan:} A kinematically feasible, collision-free trajectory to the handover pose must exist.
\item \textbf{Reach:} The object must be delivered near the receiver's hand.
\item \textbf{Stability:} The gripper must physically close on the object without dropping it.
\item \textbf{Affordance:} The robot's grasp must not occupy the receiver's intended grasp region.
\item \textbf{Safe:} The delivery motion must complete without contacting the human hand.
\end{itemize}

\subsection{Simulation Environment}\label{sec:sim_env}

All experiments are conducted in NVIDIA Isaac Sim 5.0.0~\cite{nvidia2023isaacsim}, which provides high-fidelity simulation quality and ease of use compared with engines such as MuJoCo and PyBullet.
The scene comprises a UR5e manipulator with a Robotiq 2F-85 parallel gripper, a static table, and a static MANO hand mesh representing the receiver.
Each object mesh is loaded as a rigid body with convex hull collision geometry, aligned with its canonical coordinate frame.

In the preparation stage, the benchmark provides ground-truth object representations for each trial (mesh, point cloud, voxel grid, and receiver MANO hand pose); each method reads only the inputs it requires and estimates the two poses $T_{ee}^{g}$ and $T_{ee}^{h}$ defined in Sec.~\ref{sec:task}.
In the execution stage, given $T_{ee}^{g}$, the gripper closes on the object and checks width feasibility ($w_{\mathcal{O}}(T_{ee}^{g}) \le w_{\max}$); if this check passes, the object is rigidly attached to the end-effector.
Given $T_{ee}^{h}$, the robot solves $f(q_h) = T_{ee}^{h}$ for a collision-free joint configuration and executes the planned trajectory in simulation.

\subsection{Datasets}

\begin{figure}[!t]
    \centering
    \includegraphics[width=\linewidth]{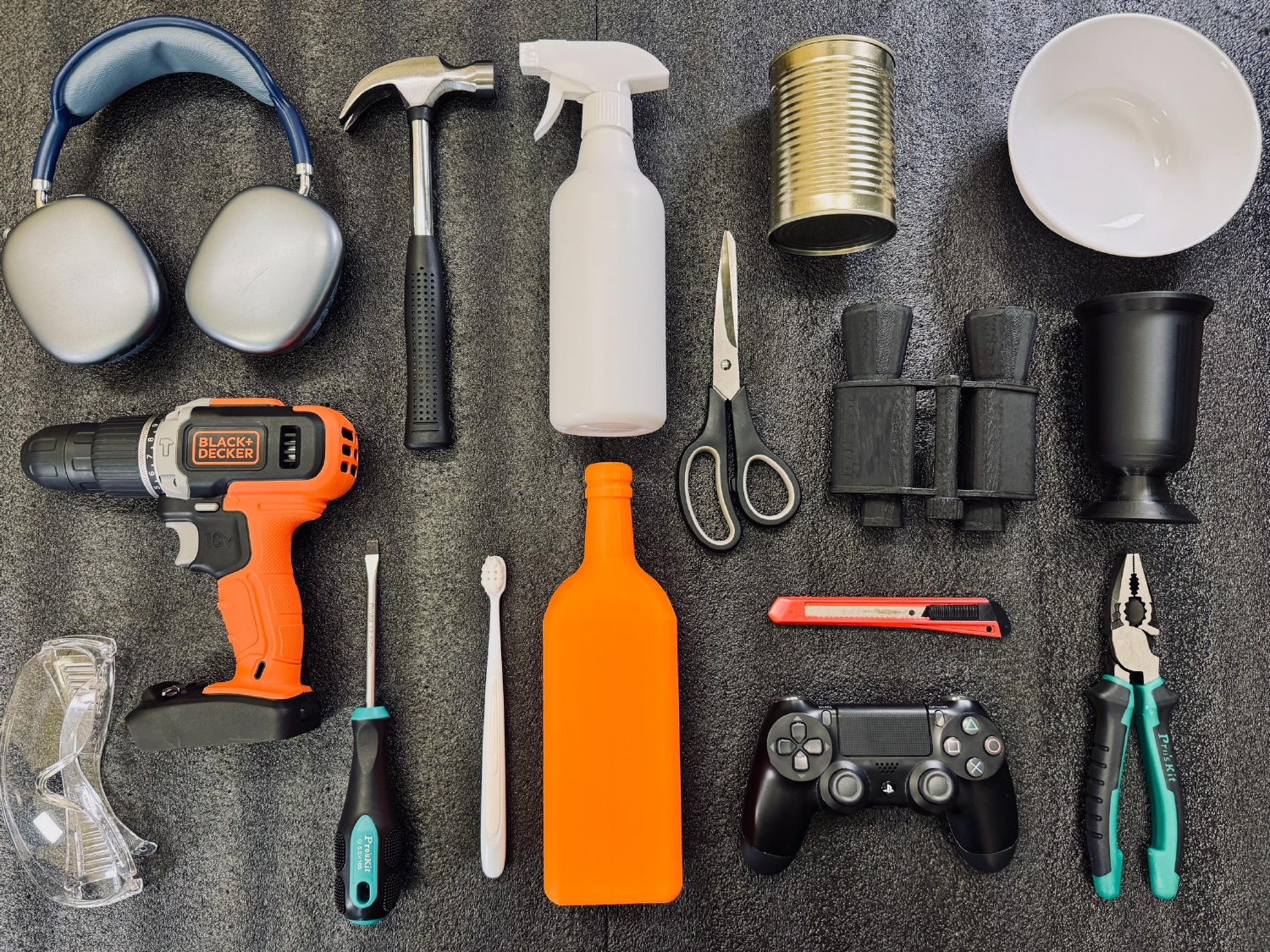}
    \caption{The 16 benchmark objects from OakInk, including objects with a functional receiver region (e.g., screwdrivers, drills), those without (e.g., bowls, game controllers), and varied sizes (cups to dispensers).}
    \label{fig:benchmark_objects}
    \vspace{-10pt}
\end{figure}

We source 16 benchmark objects from the OakInk object set~\cite{yang2022oakink}, which collectively span ShapeNet~\cite{chang2015shapenet}, YCB~\cite{calliYcbObjectModel2015}, and ContactDB~\cite{brahmbhatt2019contactdb}.
The selected objects are shown in Fig.~\ref{fig:benchmark_objects}.
OakInk is widely used in robot manipulation and hand-object interaction, enabling direct comparability with prior work.
Objects are selected to cover (1) whether the handover pose is functionally constrained (e.g., screwdrivers, drills) or admits flexible orientations (e.g., bowls, game controllers), and (2) a range of sizes from compact (e.g., cups, cans) to bulky (e.g., bottles, dispensers).

To support methods that predict hand-object interactions, we draw handover segments from H2O~\cite{ye2021h2o}, GRAB~\cite{taheriGRABDatasetWholebody2020}, and ARCTIC~\cite{fan2023arctic}, and annotate them with textual instructions (e.g., \textit{hand over the screwdriver handle to human}).
We use 200 left-hand sequences and 200 right-hand sequences, with GRAB contributing the largest share.
For robot grasp supervision, we use Multi-GraspLLM~\cite{liMultiGraspLLMMultimodalLLM2024} to generate affordance-aware grasp proposals for parallel grippers in the canonical object frame, together with functional region segmentation of the object surface.
For each object, we retain the top 100 candidate grasps.

\begin{table*}[!t]
\caption{Baseline comparison on R2HandoverSim under S0 and S1 (see Sec.~\ref{sec:eval_settings}), and Avg. Metric definitions follow Sec.~\ref{sec:metrics}.}
\vspace{-2pt}
\centering
\normalsize
\renewcommand{\arraystretch}{1.12}
\setlength{\tabcolsep}{0pt}
\begin{tabular}{@{}>{\centering\arraybackslash}m{0.048\textwidth}
>{\centering\arraybackslash}m{0.188\textwidth}
>{\columncolor{SageGreenBG}[1pt][1pt]\centering\arraybackslash}m{0.126\textwidth}
>{\columncolor{MistBlueBG}[1pt][1pt]\centering\arraybackslash}m{0.083\textwidth}
>{\columncolor{MistBlueBG}[1pt][1pt]\centering\arraybackslash}m{0.083\textwidth}
>{\columncolor{MistBlueBG}[1pt][1pt]\centering\arraybackslash}m{0.083\textwidth}
>{\columncolor{ApricotBG}[1pt][1pt]\centering\arraybackslash}m{0.077\textwidth}
>{\columncolor{ApricotBG}[1pt][1pt]\centering\arraybackslash}m{0.077\textwidth}
>{\columncolor{ApricotBG}[1pt][1pt]\centering\arraybackslash}m{0.077\textwidth}
>{\columncolor{ApricotBG}[1pt][1pt]\centering\arraybackslash}m{0.077\textwidth}
>{\columncolor{ApricotBG}[1pt][1pt]\centering\arraybackslash}m{0.077\textwidth}@{}}
\toprule
\multirow{2}{*}{\textbf{Split}}
& \multirow{2}{*}{\textbf{Methods}}
& \textbf{Success\textsuperscript{$\dagger$}~(\%)}
& \multicolumn{3}{>{\columncolor{MistBlueBG}[1pt][1pt]}c}{\textbf{Time\textsuperscript{$\dagger$} (s)}}
& \multicolumn{5}{>{\columncolor{ApricotBG}[1pt][1pt]}c}{\textbf{Failure\textsuperscript{$\dagger$} (\%)}} \\
\cmidrule(lr){3-11}
& & $\mathrm{SR}$
& $T_{\text{plan}}$
& $T_{\text{exec}}$
& $T_{\text{tot}}$
& $F_{\text{plan}}$
& $F_{\text{reach}}$
& $F_{\text{safe}}$
& $F_{\text{stab}}$
& $F_{\text{afford}}$ \\
\midrule
%%% ===============  S0  =============== %%%
\multirow{4}{*}{\rotatebox[origin=c]{90}{\textbf{S0}}}
& FC-Handover &
70.4 & 1.48 & 8.56 & 10.04 & 18.0 & 2.0 & 5.4 & 4.2 & N/A \\
& Handover-VA &
64.8 & 0.58 & 8.40 & 8.98 & 20.0 & 5.0 & 7.8 & 2.4 & N/A \\
& Contact-Handover &
77.0 & 0.92 & 8.48 & 9.40 & 13.0 & 3.6 & 2.2 & 4.2 & N/A \\
& Intent-Handover &
69.8 & 2.32 & 8.70 & 11.02 & 16.4 & 4.8 & 5.4 & 3.6 & N/A \\
\midrule\midrule
%%% ===============  S1  =============== %%%
\multirow{4}{*}{\rotatebox[origin=c]{90}{\textbf{S1}}}
& FC-Handover &
48.8 & 1.76 & 8.92 & 10.68 & 23.0 & 4.6 & 7.8 & 6.6 & 9.2 \\
& Handover-VA &
39.0 & 0.78 & 8.70 & 9.48 & 29.0 & 9.2 & 12.6 & 4.0 & 6.2 \\
& Contact-Handover &
55.4 & 1.16 & 8.76 & 9.92 & 19.6 & 8.4 & 5.0 & 7.4 & 4.2 \\
& Intent-Handover &
66.0 & 2.62 & 8.96 & 11.58 & 14.8 & 4.0 & 7.2 & 4.2 & 3.8 \\
\midrule\midrule
%%% ===============  Avg  =============== %%%
\multirow{4}{*}{\rotatebox[origin=c]{90}{\textbf{Avg}}}
& FC-Handover &
59.6 & 1.62 & 8.74 & 10.36 & 20.5 & \cellcolor{yellow!55}\textbf{3.3} & 6.6 & 5.4 & 4.6 \\
& Handover-VA &
51.9 & \cellcolor{yellow!55}\textbf{0.68} & \cellcolor{yellow!55}\textbf{8.55} & \cellcolor{yellow!55}\textbf{9.23} & 24.5 & 7.1 & 10.2 & \cellcolor{yellow!55}\textbf{3.2} & 3.1 \\
& Contact-Handover &
66.2 & 1.04 & 8.62 & 9.66 & 16.3 & 6.0 & \cellcolor{yellow!55}\textbf{3.6} & 5.8 & 2.1 \\
& Intent-Handover &
\cellcolor{yellow!55}\textbf{67.9} & 2.47 & 8.83 & 11.30 & \cellcolor{yellow!55}\textbf{15.6} & 4.4 & 6.3 & 3.9 & \cellcolor{yellow!55}\textbf{1.9} \\
\bottomrule
\multicolumn{11}{@{}l}{\footnotesize\textsuperscript{$\dagger$}Averaged over all trials per object per method. For $F_{\text{afford}}$, S0 is not evaluated; in Avg, S0 is treated as 0 and averaged with S1.} \\
\multicolumn{11}{@{}l}{\footnotesize Highlighting in Avg indicates best values per metric (higher for $\mathrm{SR}$; lower for time and failure rates).} \\
\end{tabular}
\vspace{-10pt}
\label{tab:baseline_results}
\end{table*}

\subsection{Evaluation Metrics}\label{sec:metrics}

We evaluate each trial across five binary metrics: planning feasibility, reachability, grasp stability, grasp affordance, and safety. Throughout this subsection, let $g = T_{ee}^g$ and $h = T_{ee}^h$ for conciseness.

Planning feasibility checks whether the robot can reach handover pose $h$ with a collision-free IK solution and a feasible motion plan under grasp $g$:
\begin{equation}
\text{Plan}(g, h)
\;\Leftrightarrow\;
h \in \mathcal{S}_{\text{plan}}(g),
\end{equation}

where $\mathcal{S}_{\text{plan}}(g) \subset SE(3)$ denotes the set of handover poses that satisfy both conditions.

Reachability evaluates whether the object is delivered into a predefined region near the human hand.
Let $\mathcal{B}_h$ denote a sphere centered at $\mathbf{c}_h = \mathbf{p}_h + d_h \mathbf{n}_h$ with radius $r_h$, where $\mathbf{p}_h$ and $\mathbf{n}_h$ are the palm position and outward normal for each trial; we set $d_h=12$\,cm and $r_h=10$\,cm.
Let $\mathcal{V}_{\mathcal{O}}(g, h)$ denote the object volume in the world frame when the end effector is at $h$ under grasp $g$.
Reachability is defined as
\begin{equation}
\text{Reach}(g, h)
\;\Leftrightarrow\;
\mathcal{V}_{\mathcal{O}}(g, h) \cap \mathcal{B}_h \neq \emptyset.
\end{equation}

Grasp stability evaluates whether the gripper can retain the object at the predicted grasp pose.
Given $g$, let $w_{\mathcal{O}}(g)$ denote the object width along the gripper closing axis, and let $w_{\max}=85$\,mm denote the maximum aperture of the Robotiq 2F-85 gripper.
Stability is defined as
\begin{equation}
\text{Stability}(g)
\;\Leftrightarrow\;
w_{\mathcal{O}}(g) \le w_{\max}.
\end{equation}

Grasp affordance evaluates whether the robot fingers occlude the receiver's intended grasp area.
For each object $\mathcal{O}$, let $\mathcal{G}_{\mathcal{O}}^{H} \subset \mathbb{R}^3$ denote the grasp region intended for the receiver, and let $\mathcal{G}_{\mathcal{O}}^{H}(g)$ be its rigid transform in the world frame under $g$.
Let $\mathcal{V}_{\text{grip}}(g)$ denote the volume occupied by the gripper fingers at pose $g$.
Affordance is defined as
\begin{equation}
\text{Affordance}(g)
\;\Leftrightarrow\;
\mathcal{V}_{\text{grip}}(g) \cap \mathcal{G}_{\mathcal{O}}^{H}(g) = \emptyset.
\end{equation}

Safety evaluates whether the human hand contacts the robot arm or end effector during execution.
Let $\mathcal{M}_{\text{robot}}(g, h)$ denote the collision volume of the robot arm and end effector during trajectory execution under $(g,h)$, and let $\mathcal{M}_{\text{hand}}$ denote the static MANO hand mesh.
Safety is defined as
\begin{equation}
\text{Safe}(g, h)
\;\Leftrightarrow\;
\mathcal{M}_{\text{hand}} \cap \mathcal{M}_{\text{robot}}(g, h) = \emptyset.
\end{equation}

We report $\mathrm{SR}$ (success rate), $T_{\text{plan}}$ (planning time), $T_{\text{exec}}$ (execution time), $T_{\text{tot}}=T_{\text{plan}}+T_{\text{exec}}$ (total time), and failure rates $F_{\text{plan}}$, $F_{\text{reach}}$, $F_{\text{safe}}$, $F_{\text{stab}}$, and $F_{\text{afford}}$.
In S1 (see Sec.~\ref{sec:eval_settings}), the five criteria are evaluated sequentially: Stability is checked first (gripper closure), then Plan (IK and motion planning), then Reach, Affordance, and Safe.
In S0 (see Sec.~\ref{sec:eval_settings}), the Affordance check is not evaluated. Therefore, $F_{\text{afford}}$ is not reported for S0 and does not contribute to $\mathrm{SR}$ in S0.
A trial is attributed to the \emph{first} evaluated criterion that fails; consequently, the reported failure rates are mutually exclusive and sum to $1-\mathrm{SR}$ within each split.
The overall outcome of a trial is defined as
\begin{equation}
\text{Success}(g, h)
=
\left\{
\begin{aligned}
& h \in \mathcal{S}_{\text{plan}}(g) \\[2pt]
& \mathcal{V}_{\mathcal{O}}(g, h) \cap \mathcal{B}_h \neq \emptyset \\[2pt]
& w_{\mathcal{O}}(g) \le w_{\max} \\[2pt]
& \mathcal{V}_{\text{grip}}(g) \cap \mathcal{G}_{\mathcal{O}}^{H}(g) = \emptyset \\[2pt]
& \mathcal{M}_{\text{hand}} \cap \mathcal{M}_{\text{robot}}(g, h) = \emptyset
\end{aligned}
\right.
\label{eq:success}
\end{equation}

where Eq.~\eqref{eq:success} is the full criterion set used in S1, and S0 uses the same definition without the Affordance term.

\begin{figure*}[!t]
    \centering
    \begin{subfigure}[t]{\textwidth}
        \centering
        \includegraphics[width=\linewidth]{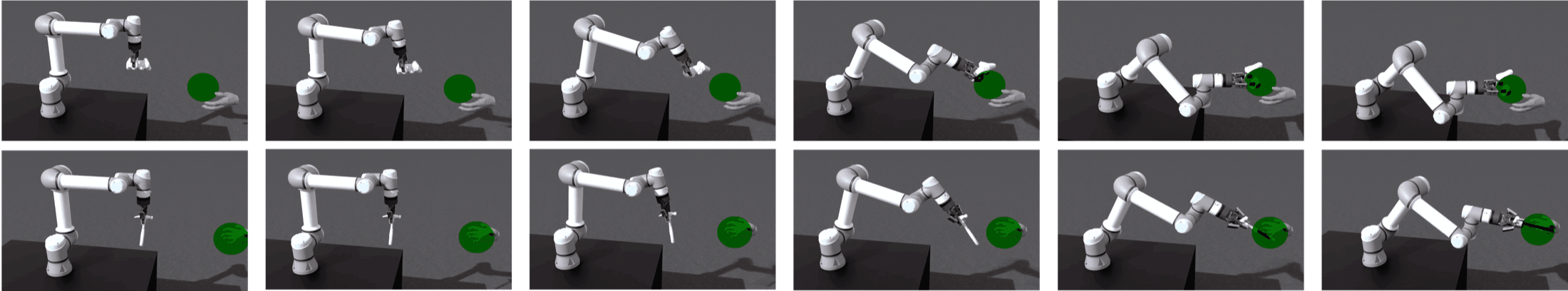}
        \caption{FC-Handover}
        \label{fig:fc_handover}
    \end{subfigure}\par
    
    \vspace{1mm}
    
    \begin{subfigure}[t]{\textwidth}
        \centering
        \includegraphics[width=\linewidth]{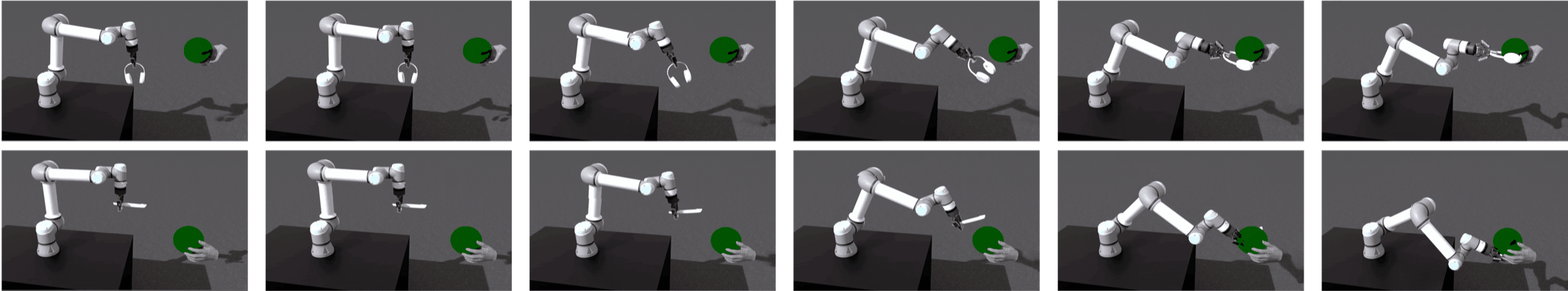}
        \caption{Handover-VA}
        \label{fig:handover_va}
    \end{subfigure}\par
    
    \vspace{1mm}
    
    \begin{subfigure}[t]{\textwidth}
        \centering
        \includegraphics[width=\linewidth]{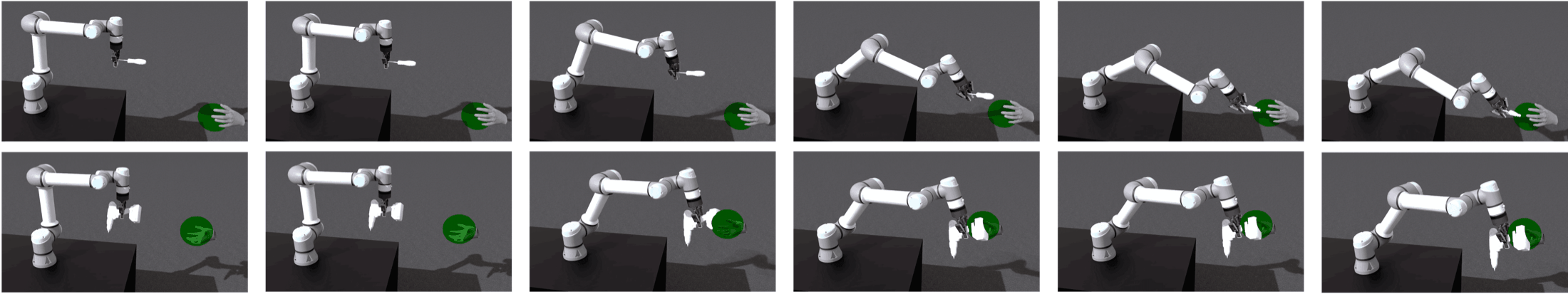}
        \caption{Contact-Handover}
        \label{fig:contact_handover}
    \end{subfigure}\par

    \vspace{1mm}

    \begin{subfigure}[t]{\textwidth}
        \centering
        \includegraphics[width=\linewidth]{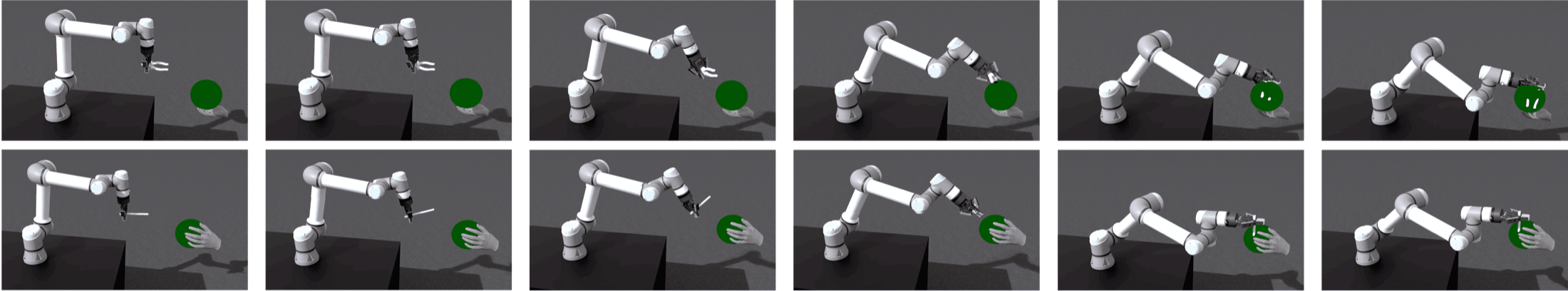}
        \caption{Intent-Handover}
        \label{fig:intent_handover}
    \end{subfigure}
    \caption{Qualitative comparison of predicted handover configurations across four baselines. Each row shows the grasp pose, approach trajectory, and final handover pose for the same object--hand pair.}
    \label{fig:qual_comparison}
    \vspace{-10pt}
\end{figure*}

\subsection{Evaluation Settings}\label{sec:eval_settings}

We evaluate all methods under two splits: S0 and S1.
S0 contains objects with relatively unconstrained handover configurations.
S1 contains objects with stronger functional constraints on grasp and handover orientation.
Accordingly, affordance failure $F_{\text{afford}}$ is evaluated in S1 but not in S0.
As a result, $\mathrm{SR}$ in S0 is computed without the affordance criterion.
These two splits correspond to the S0 and S1 rows in Table~\ref{tab:baseline_results}, and Avg reports their average performance.

The robot does not perform a tabletop pick-up; instead, the gripper is initialized vertically downward and the object is loaded at the generated grasp pose relative to the end-effector (see Sec.~\ref{sec:task}).
This broadens the feasible grasp range for each object and isolates handover performance from artifacts of the simulated grasping process.

\section{Experiments}

\subsection{Baseline Implementation}

\textbf{FC-Handover \cite{meng2022fast}.}
This method uses GanHand \cite{coronaGanhandPredictingHuman2020} to predict the human receiving pose.
GraspIt! \cite{millerGraspitVersatileSimulator2004} then generates robot grasp candidates opposite the human hand to avoid collision.
The robot gripper is aligned 180$^\circ$ opposite the target hand orientation, calculated from the wrist and middle finger tip positions.

\textbf{Handover-VA \cite{lehotsky2023optimizing}.}
This method uses AffNet-DR \cite{christensenLearningSegmentObject2022} to partition the object into functional regions, within which ICP determines the grasp pose.
While the original method targets the midpoint between human and robot, we adapt this by placing the gripper opposite the target hand's palm center.
The orientation follows the original rule-based strategy, aligning the object's utility axis to present the handle to the human.

\textbf{Contact-Handover \cite{wang2024contacthandover}.}
This system uses 3D VoxNet \cite{maturanaVoxnet3dConvolutional2015} to predict contact heatmaps and Contact-GraspNet \cite{sundermeyerContactgraspnetEfficient6dof2021} for grasp candidates.
DBSCAN clusters grasps and filters collisions with the MANO hand to rerank the optimal pose.
To mitigate computational costs, we perform offline generation using \textit{Open3D}'s \textit{RaycastingScene} for efficient batch collision detection.
The handover target is set 15 cm above the estimated palm center, with orientation minimizing collision cost.

\textbf{Intent-Handover}.
This method uses Text2HOI \cite{cha2024text2hoi} to generate hand-object interaction poses.
Grasp candidates are evaluated by a joint affordance and safety score, and the highest-scoring candidate is selected as the optimal grasp.
We adopt the same handover position strategy as Contact-Handover.
The orientation follows the original method: a 15$^\circ$ inward rotation along the predicted palm plane.

\subsection{Results}

Table~\ref{tab:baseline_results} reports baseline performance under S0 (unconstrained objects), S1 (functionally constrained objects), and their average.
Qualitative examples are shown in Fig.~\ref{fig:qual_comparison}.

\textbf{Contact-Handover} achieves the highest S0 success rate (77.0\%) and the lowest average $F_{\text{safe}}$ (3.6\%).
Under S1, all kinematic failure modes worsen: $F_{\text{plan}}$ rises from 13.0\% to 19.6\%, $F_{\text{reach}}$ from 3.6\% to 8.4\%, $F_{\text{safe}}$ from 2.2\% to 5.0\%, and $F_{\text{stab}}$ from 4.2\% to 7.4\%, indicating that the collision-cost objective does not transfer to functional constraints.
In S1, $F_{\text{afford}}$ is 4.2\% (second lowest), suggesting that contact priors from human interaction data implicitly encode functional grasp preferences.

\textbf{Intent-Handover} achieves the highest average success rate (67.9\%) and the highest S1 success rate (66.0\%).
The lowest average $F_{\text{plan}}$ (15.6\%) and S1 $F_{\text{afford}}$ (3.8\%) indicate that language-conditioned pose generation steers candidates toward configurations that satisfy functional constraints.
$T_{\text{plan}}$ averages 2.47\,s (2.4$\times$ Contact-Handover), and $F_{\text{safe}}$ reaches 5.4\% in S0, as collision avoidance is applied against a generated hand pose rather than a directly estimated one.

\textbf{FC-Handover} (Avg SR: 59.6\%) records the lowest average $F_{\text{reach}}$ (3.3\%), with $F_{\text{reach}}$ rising only from 2.0\% to 4.6\% across splits, indicating that the placement opposite the predicted hand produces poses within the robot kinematic workspace.
In S1, $F_{\text{afford}}$ reaches 9.2\% (highest) and $F_{\text{plan}}$ rises to 23.0\%, showing the orientation strategy does not encode which object region should face the receiver.

\textbf{Handover-VA} records the lowest average SR (51.9\%) but the shortest $T_{\text{plan}}$ (0.68\,s) and lowest $F_{\text{stab}}$ (3.2\%).
The highest average $F_{\text{plan}}$ (24.5\%) and $F_{\text{safe}}$ (10.2\%), together with the largest S0-to-S1 drop ($-$25.8\,pp), indicate that the orientation heuristic does not account for robot kinematics or collision geometry.

\begin{figure}[!t]
    \centering
    \includegraphics[width=\columnwidth]{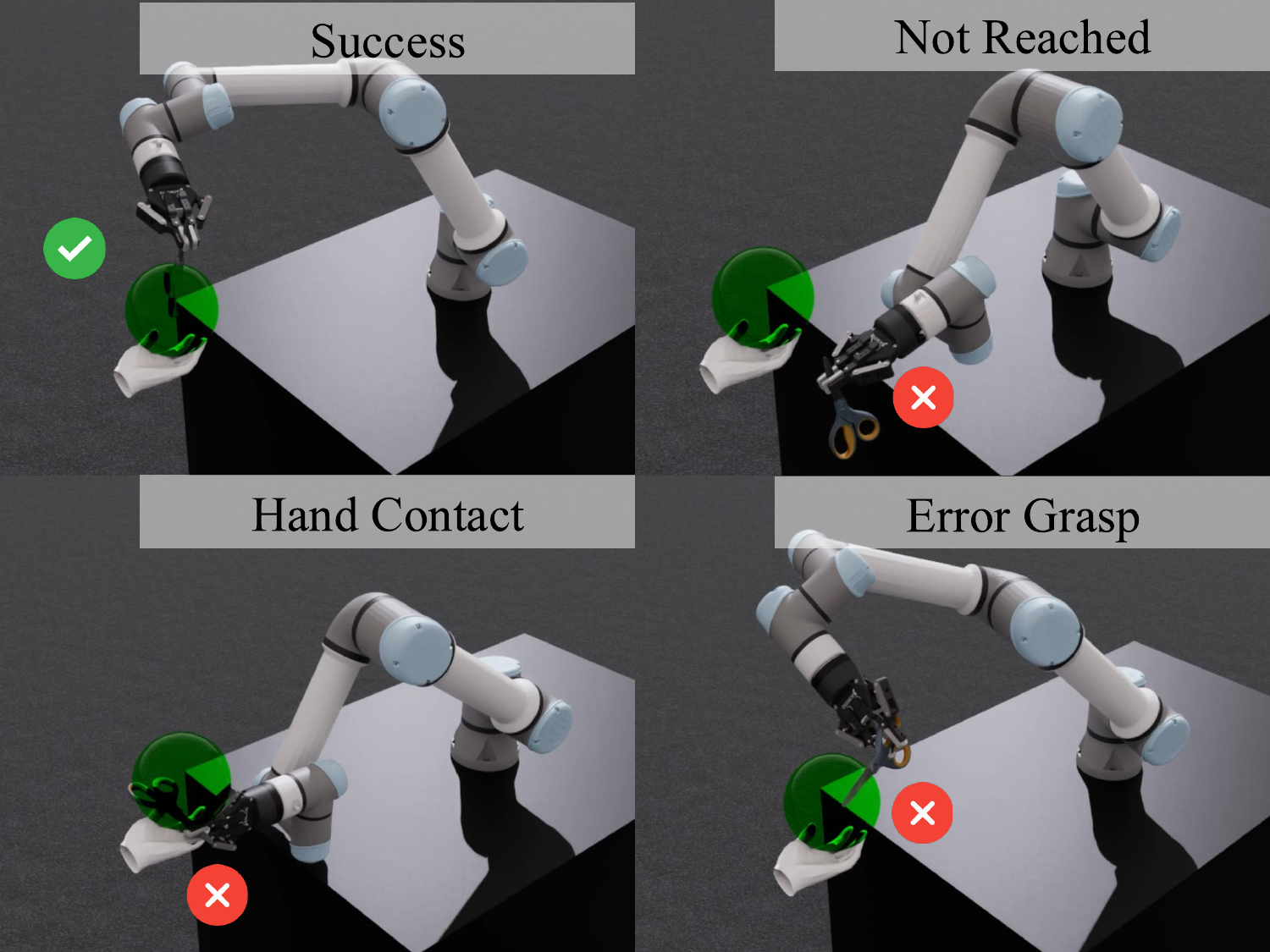}
    \caption{Representative success and failure cases in simulation. Failures are categorized by planning infeasibility, reachability violation, hand--gripper collision, object drop, and functional-region occlusion.}
    \label{fig:failure_cases}
    \vspace{-10pt}
\end{figure}

\textbf{Cross-method analysis.}
Intent-Handover and Contact-Handover rank first and second in S1 $F_{\text{afford}}$ (3.8\% and 4.2\%), while FC-Handover and Handover-VA reach 9.2\% and 6.2\%, confirming that functional awareness (via language or contact priors) reduces affordance failure under constrained objects.
Each remaining method leads in one metric: FC-Handover in $F_{\text{reach}}$ (3.3\%), Contact-Handover in $F_{\text{safe}}$ (3.6\%), and Handover-VA in $F_{\text{stab}}$ (3.2\%), while Intent-Handover incurs the highest $T_{\text{plan}}$ (2.47\,s).
These complementary strengths suggest that combining collision-cost ranking with language-guided affordance generation could address the main failure modes in both settings.
Fig.~\ref{fig:failure_cases} illustrates representative success and failure cases.

\subsection{Sim-to-Real Experiment}

\begin{figure}[!t]
    \centering
    \includegraphics[width=\columnwidth]{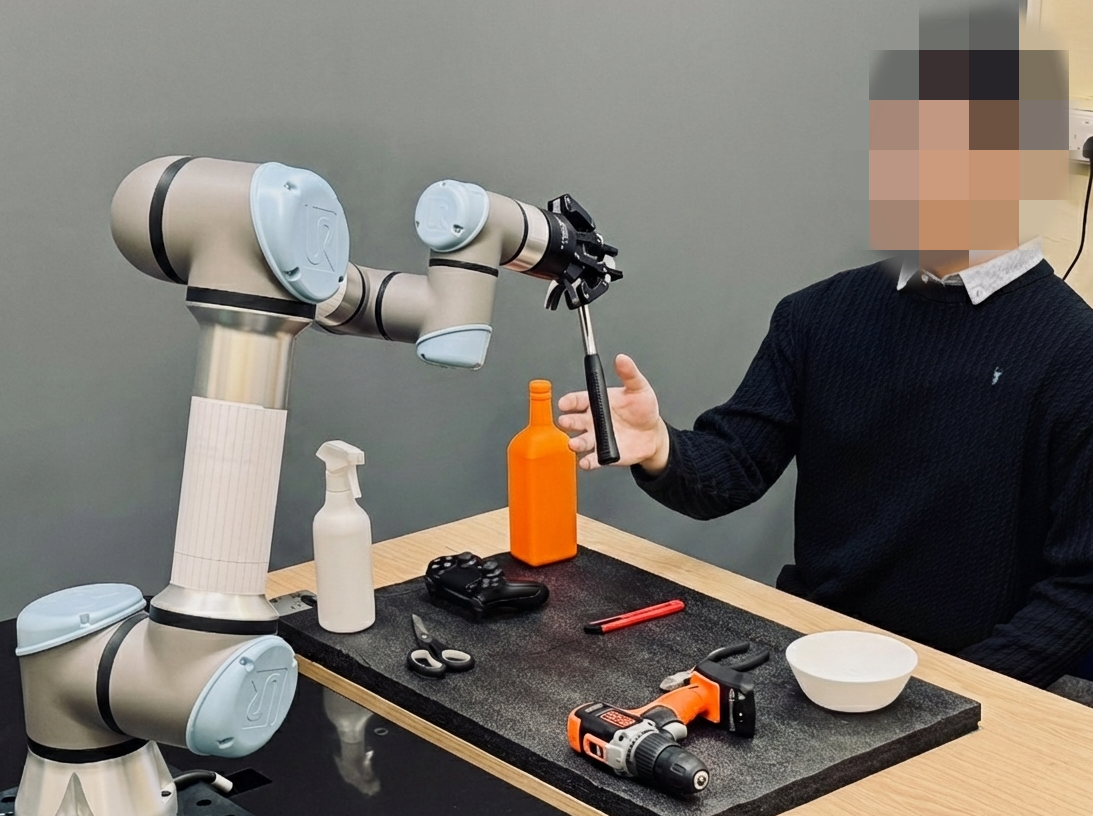}
    \caption{Real hardware setup for the sim-to-real experiment. A UR5e robot with a Robotiq 2F-85 gripper delivers objects to a seated human receiver.}
    \label{fig:real_hardware_setup}
\end{figure}

We deploy all four baselines on the physical platform (Fig.~\ref{fig:real_hardware_setup}) and evaluate each method with 30 participants (5~trials per method per participant, 600~total trials).
For each participant, five objects are randomly drawn: 2 from S0 and 3 from S1.
The same five objects are used across all four methods (one trial per object), and method order is randomized to support within-subject comparison.
For objects whose geometry prevents stable tabletop resting (e.g., knife, game controller), a 3D-printed fixture supports the object and exposes the designated grasp region, allowing the robot arm to approach from a wider range of angles.

Table~\ref{tab:sim2real} reports success rates and total times.
Intent-Handover achieves the highest real-world success rate (73.3\%, $\Delta$SR\,=\,+5.4\,pp),
followed by FC-Handover (63.3\%, $\Delta$SR\,=\,+3.7\,pp).
Handover-VA shows the largest success rate gain (60.0\%, $\Delta$SR\,=\,+8.1\,pp).
Contact-Handover is the only method with negative transfer ($\Delta$SR\,=\,$-$9.5\,pp, from 66.2\% to 56.7\%).
We attribute this drop to the sensitivity of its voxel-based contact predictions to real-world sensor noise and extrinsic calibration errors, which can partially occlude the receiver grasp region.
Contact-Handover was also validated exclusively in simulation, without real-world adaptation.

All methods exhibit increased total time in the real world ($\Delta T$\,=\,3.12--3.48\,s), because $T_{\text{tot}}$ in simulation measures only robot planning and execution, whereas the real world additionally includes human reception time (visual assessment, wrist reorientation, and grasp closure).

\begin{table}[h]
\caption{Sim-to-real transfer results. $\Delta$ is computed as real-world result minus simulation Avg in Table~\ref*{tab:baseline_results}.}
\vspace{-2pt}
\centering
\normalsize
\renewcommand{\arraystretch}{1.12}
\setlength{\tabcolsep}{0pt}
\begin{tabular}{@{}>{\raggedright\arraybackslash}m{0.31\columnwidth}
>{\columncolor{SageGreenBG}\centering\arraybackslash}m{0.19\columnwidth}
>{\columncolor{SageGreenBG}\centering\arraybackslash}m{0.12\columnwidth}
>{\columncolor{MistBlueBG}\centering\arraybackslash}m{0.23\columnwidth}
>{\columncolor{MistBlueBG}\centering\arraybackslash}m{0.15\columnwidth}@{}}
\toprule
\textbf{Methods}
& \mbox{\textbf{SR\textsuperscript{$\dagger$} (\%)}}
& $\boldsymbol{\Delta}$
& \mbox{\textbf{$T_{\text{tot}}$\textsuperscript{$\dagger$} (s)}}
& $\boldsymbol{\Delta}$ \\
\midrule
\mbox{FC-Handover} & 63.3 & +3.7 & 13.52 & +3.16 \\
\mbox{Handover-VA} & 60.0 & \cellcolor{orange!55}\textbf{+8.1} & \cellcolor{yellow!55}\textbf{12.35} & \multicolumn{1}{>{\columncolor{orange!55}\centering\arraybackslash}m{0.15\columnwidth}@{}}{\textbf{+3.12}} \\
\mbox{Contact-Handover} & 56.7 & $-$9.5 & 13.14 & +3.48 \\
\mbox{Intent-Handover} & \cellcolor{yellow!55}\textbf{73.3} & +5.4 & 14.62 & +3.32 \\
\bottomrule
\multicolumn{5}{@{}p{\columnwidth}@{}}{\footnotesize\textsuperscript{$\dagger$}Averaged over all trials.} \\
\multicolumn{5}{@{}p{\columnwidth}@{}}{\footnotesize Highlighting: Yellow indicates best real-world performance (higher SR, lower Time). Orange indicates largest SR increase and smallest Time increase.} \\
\end{tabular}
\vspace{-10pt}
\label{tab:sim2real}
\end{table}

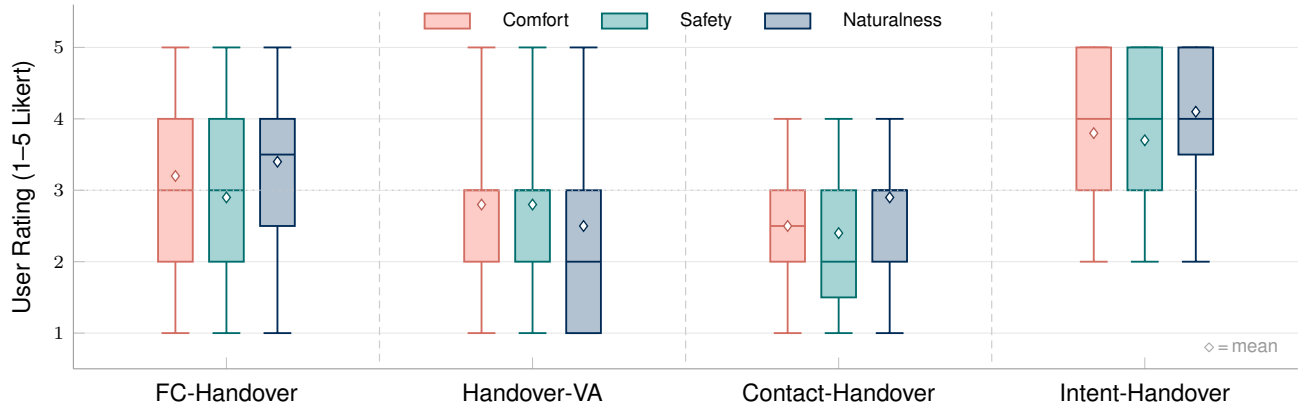
\begin{figure*}[!t]
\centering
\begin{tikzpicture}
\begin{axis}[
  width=\textwidth,
  height=6.4cm,
  boxplot/draw direction=y,
  ylabel={User Rating (1--5 Likert)},
  ymin=0.5, ymax=5.6,
  ytick={1,2,3,4,5},
  % Group centers: FC=2.0  VA=5.0  Contact=8.0  Intent=11.0
  % Within-group: Comfort=-0.5  Safety=0  Naturalness=+0.5
  xtick={2.0, 5.0, 8.0, 11.0},
  xticklabels={FC-Handover, Handover-VA, Contact-Handover, Intent-Handover},
  x tick label style={font=\small\sffamily, anchor=north, yshift=-2pt},
  ylabel style={font=\small\sffamily},
  y tick label style={font=\scriptsize},
  % --- Nature / Science Robotics axis style ---
  ymajorgrids=true,
  grid style={line width=0.15pt, draw=black!10},
  axis line style={draw=black!40, line width=0.4pt},
  tick style={draw=black!25, line width=0.3pt},
  axis x line*=bottom,
  axis y line*=left,
  clip=false,
  xmin=0.5, xmax=12.5,
  legend style={
  at={(0.5, 0.90)}, anchor=south, % Pull legend closer to boxplots
  font=\scriptsize\sffamily,
  draw=none, fill=none,
  legend columns=3,
  column sep=10pt,
  },
  legend columns=3,
  legend cell align={left},
]

% ======================================================================
%  COLORS:  Comfort=natureSalmon   Safety=natureTeal   Naturalness=natureBlue
%
%  All \addplot calls use [forget plot] to suppress auto-legend entries.
%  Legend is built manually via \addlegendimage below.
%
%  Data (n=30).  Quartiles derived from sorted distributions.
%  Q1 = 8th value, Median = avg(15th,16th), Q3 = 23rd value (n=30).
% ======================================================================

% ----------------------------------------------------------------
%  FC-Handover  (group center x=2.0)
% ----------------------------------------------------------------
% Comfort  x=1.5 | {1x2,2x6,3x10,4x8,5x4} | Q1=2 med=3 Q3=4 | mean=3.20
\addplot[forget plot,
  boxplot prepared={draw position=1.5, box extend=0.34,
    lower whisker=1, lower quartile=2, median=3, upper quartile=4, upper whisker=5},
  color=natureSalmon!85!black, fill=natureSalmon!40, solid, line width=0.7pt,
] coordinates {};
\addplot[forget plot, only marks, mark=diamond*, mark size=2.0pt,
  mark options={solid, fill=white, draw=natureSalmon!75!black, line width=0.5pt}]
  coordinates {(1.5, 3.20)};

% Safety   x=2.0 | {1x4,2x7,3x10,4x6,5x3} | Q1=2 med=3 Q3=4 | mean=2.90
\addplot[forget plot,
  boxplot prepared={draw position=2.0, box extend=0.34,
    lower whisker=1, lower quartile=2, median=3, upper quartile=4, upper whisker=5},
  color=natureTeal!85!black, fill=natureTeal!35, solid, line width=0.7pt,
] coordinates {};
\addplot[forget plot, only marks, mark=diamond*, mark size=2.0pt,
  mark options={solid, fill=white, draw=natureTeal!75!black, line width=0.5pt}]
  coordinates {(2.0, 2.90)};

% Naturalness x=2.5 | {1x2,2x5,3x8,4x9,5x6} | Q1=2.5 med=3.5 Q3=4 | mean=3.40
\addplot[forget plot,
  boxplot prepared={draw position=2.5, box extend=0.34,
    lower whisker=1, lower quartile=2.5, median=3.5, upper quartile=4, upper whisker=5},
  color=natureBlue!85!black, fill=natureBlue!30, solid, line width=0.7pt,
] coordinates {};
\addplot[forget plot, only marks, mark=diamond*, mark size=2.0pt,
  mark options={solid, fill=white, draw=natureBlue!75!black, line width=0.5pt}]
  coordinates {(2.5, 3.40)};

% ----------------------------------------------------------------
%  Handover-VA  (group center x=5.0)
% ----------------------------------------------------------------
% Comfort  x=4.5 | {1x4,2x6,3x14,4x4,5x2} | Q1=2 med=3 Q3=3 | mean=2.80
\addplot[forget plot,
  boxplot prepared={draw position=4.5, box extend=0.34,
    lower whisker=1, lower quartile=2, median=3, upper quartile=3, upper whisker=5},
  color=natureSalmon!85!black, fill=natureSalmon!40, solid, line width=0.7pt,
] coordinates {};
\addplot[forget plot, only marks, mark=diamond*, mark size=2.0pt,
  mark options={solid, fill=white, draw=natureSalmon!75!black, line width=0.5pt}]
  coordinates {(4.5, 2.80)};

% Safety   x=5.0 | {1x4,2x7,3x12,4x5,5x2} | Q1=2 med=3 Q3=3 | mean=2.80
\addplot[forget plot,
  boxplot prepared={draw position=5.0, box extend=0.34,
    lower whisker=1, lower quartile=2, median=3, upper quartile=3, upper whisker=5},
  color=natureTeal!85!black, fill=natureTeal!35, solid, line width=0.7pt,
] coordinates {};
\addplot[forget plot, only marks, mark=diamond*, mark size=2.0pt,
  mark options={solid, fill=white, draw=natureTeal!75!black, line width=0.5pt}]
  coordinates {(5.0, 2.80)};

% Naturalness x=5.5 | {1x8,2x7,3x9,4x4,5x2} | Q1=1 med=2 Q3=3 | mean=2.50
%  (fixed orientation heuristic -> lowest naturalness)
\addplot[forget plot,
  boxplot prepared={draw position=5.5, box extend=0.34,
    lower whisker=1, lower quartile=1, median=2, upper quartile=3, upper whisker=5},
  color=natureBlue!85!black, fill=natureBlue!30, solid, line width=0.7pt,
] coordinates {};
\addplot[forget plot, only marks, mark=diamond*, mark size=2.0pt,
  mark options={solid, fill=white, draw=natureBlue!75!black, line width=0.5pt}]
  coordinates {(5.5, 2.50)};

% ----------------------------------------------------------------
%  Contact-Handover  (group center x=8.0)
% ----------------------------------------------------------------
% Comfort  x=7.5 | {1x5,2x10,3x10,4x5} | Q1=2 med=2.5 Q3=3 | mean=2.50
\addplot[forget plot,
  boxplot prepared={draw position=7.5, box extend=0.34,
    lower whisker=1, lower quartile=2, median=2.5, upper quartile=3, upper whisker=4},
  color=natureSalmon!85!black, fill=natureSalmon!40, solid, line width=0.7pt,
] coordinates {};
\addplot[forget plot, only marks, mark=diamond*, mark size=2.0pt,
  mark options={solid, fill=white, draw=natureSalmon!75!black, line width=0.5pt}]
  coordinates {(7.5, 2.50)};

% Safety   x=8.0 | {1x6,2x10,3x10,4x4} | Q1=1.5 med=2 Q3=3 | mean=2.40
%  (grasp occlusion forces awkward repositioning -> lowest safety)
\addplot[forget plot,
  boxplot prepared={draw position=8.0, box extend=0.34,
    lower whisker=1, lower quartile=1.5, median=2, upper quartile=3, upper whisker=4},
  color=natureTeal!85!black, fill=natureTeal!35, solid, line width=0.7pt,
] coordinates {};
\addplot[forget plot, only marks, mark=diamond*, mark size=2.0pt,
  mark options={solid, fill=white, draw=natureTeal!75!black, line width=0.5pt}]
  coordinates {(8.0, 2.40)};

% Naturalness x=8.5 | {1x2,2x8,3x11,4x9} | Q1=2 med=3 Q3=3 | mean=2.90
\addplot[forget plot,
  boxplot prepared={draw position=8.5, box extend=0.34,
    lower whisker=1, lower quartile=2, median=3, upper quartile=3, upper whisker=4},
  color=natureBlue!85!black, fill=natureBlue!30, solid, line width=0.7pt,
] coordinates {};
\addplot[forget plot, only marks, mark=diamond*, mark size=2.0pt,
  mark options={solid, fill=white, draw=natureBlue!75!black, line width=0.5pt}]
  coordinates {(8.5, 2.90)};

% ----------------------------------------------------------------
%  Intent-Handover  (group center x=11.0)
% ----------------------------------------------------------------
% Comfort  x=10.5 | {2x4,3x7,4x10,5x9} | Q1=3 med=4 Q3=5 | mean=3.80
\addplot[forget plot,
  boxplot prepared={draw position=10.5, box extend=0.34,
    lower whisker=2, lower quartile=3, median=4, upper quartile=5, upper whisker=5},
  color=natureSalmon!85!black, fill=natureSalmon!40, solid, line width=0.7pt,
] coordinates {};
\addplot[forget plot, only marks, mark=diamond*, mark size=2.0pt,
  mark options={solid, fill=white, draw=natureSalmon!75!black, line width=0.5pt}]
  coordinates {(10.5, 3.80)};

% Safety   x=11.0 | {2x4,3x8,4x11,5x7} | Q1=3 med=4 Q3=5 | mean=3.70
\addplot[forget plot,
  boxplot prepared={draw position=11.0, box extend=0.34,
    lower whisker=2, lower quartile=3, median=4, upper quartile=5, upper whisker=5},
  color=natureTeal!85!black, fill=natureTeal!35, solid, line width=0.7pt,
] coordinates {};
\addplot[forget plot, only marks, mark=diamond*, mark size=2.0pt,
  mark options={solid, fill=white, draw=natureTeal!75!black, line width=0.5pt}]
  coordinates {(11.0, 3.70)};

% Naturalness x=11.5 | {2x1,3x7,4x10,5x12} | Q1=3.5 med=4 Q3=5 | mean=4.10
%  (language model encodes task semantics -> highest naturalness)
\addplot[forget plot,
  boxplot prepared={draw position=11.5, box extend=0.34,
    lower whisker=2, lower quartile=3.5, median=4, upper quartile=5, upper whisker=5},
  color=natureBlue!85!black, fill=natureBlue!30, solid, line width=0.7pt,
] coordinates {};
\addplot[forget plot, only marks, mark=diamond*, mark size=2.0pt,
  mark options={solid, fill=white, draw=natureBlue!75!black, line width=0.5pt}]
  coordinates {(11.5, 4.10)};

% ==================== Group separators ====================
\draw[densely dashed, black!20, line width=0.45pt]
  (axis cs:3.5, 0.6) -- (axis cs:3.5, 5.5);
\draw[densely dashed, black!20, line width=0.45pt]
  (axis cs:6.5, 0.6) -- (axis cs:6.5, 5.5);
\draw[densely dashed, black!20, line width=0.45pt]
  (axis cs:9.5, 0.6) -- (axis cs:9.5, 5.5);

% Scale midpoint reference line
\draw[dotted, black!22, line width=0.5pt]
  (axis cs:0.5, 3) -- (axis cs:12.5, 3);

% ==================== Legend ====================
\addlegendimage{area legend,
  fill=natureSalmon!40, draw=natureSalmon!85!black, line width=0.7pt}
\addlegendentry{Comfort}
\addlegendimage{area legend,
  fill=natureTeal!35, draw=natureTeal!85!black, line width=0.7pt}
\addlegendentry{Safety}
\addlegendimage{area legend,
  fill=natureBlue!30, draw=natureBlue!85!black, line width=0.7pt}
\addlegendentry{Naturalness}

% Mean marker annotation
\node[anchor=south east, font=\scriptsize\sffamily, text=black!40]
  at (axis cs:12.4, 0.6) {$\diamond$\,=\,mean};

\end{axis}
\end{tikzpicture}
\caption{%
  Subjective ratings for handover comfort, safety, and naturalness ($n = 30$; Likert scale from 1 to 5).
  The boxes span the interquartile range (Q1 to Q3), the horizontal lines indicate medians, and the diamonds mark means.
  The whiskers extend to the minimum and maximum ratings.
  The dotted horizontal line marks the scale midpoint (3).%
}
\label{fig:user_ratings}
\end{figure*}

The other three baselines, each validated or calibrated on physical hardware, all achieve positive transfer.
We attribute these gains partly to active receiver adaptation: participants adjusted their wrist orientation and approach angle to accommodate the presented object, a compensatory behavior absent in the static simulation receiver.
We additionally observe that object weight in the real world increases object drop failures relative to simulation; even with the support fixture, instability during delivery remains an open challenge.

While modeling the adaptive behavior of human receivers in simulation remains an open challenge, R2HandoverSim nonetheless discriminates effectively among methods: it reveals distinct failure mode profiles across baselines (Table~\ref{tab:baseline_results}), correctly identifies the top-performing method in real-world deployment, and exposes divergent sim-to-real transfer characteristics (e.g., Contact-Handover's negative transfer versus positive transfer for the other three methods).

\subsection{User Study}\label{sec:user_study}

Simulation success rate captures task completion but not user experience: a technically successful handover may still feel uncomfortable, unsafe, or unnatural to the receiver.
We test the hypothesis that \textit{higher simulation success rate tends to predict higher perceived handover quality across comfort, perceived safety, and naturalness.}

Beyond task success, we collect subjective assessments from the same 30~participants.
After completing all trials of a given method, each participant rates the handover on three independent 5-point Likert scales: comfort, perceived safety, and naturalness (1\,=\,very poor, 5\,=\,very good).
Fig.~\ref{fig:user_ratings} summarizes the distributions.
A Friedman test reveals significant differences among the four methods for all three dimensions: comfort ($\chi^2(3)=28.4$, $p<0.001$), perceived safety ($\chi^2(3)=24.1$, $p<0.001$), and naturalness ($\chi^2(3)=31.7$, $p<0.001$).
Post-hoc Wilcoxon signed-rank tests with Bonferroni correction ($\alpha=0.05/6$) confirm that Intent-Handover is rated significantly higher than every other method across all three dimensions ($p<0.005$ in all pairwise comparisons).

Intent-Handover achieves the highest ratings across all three dimensions.
Participants rated it highest in naturalness (median\,=\,4, Q1\,=\,3.5, mean\,=\,4.1), followed by comfort (mean\,=\,3.8) and perceived safety (mean\,=\,3.7).
We attribute this to the language-guided diffusion model consistently presenting objects at orientations that match the expected grasp, reducing the wrist adjustment required before closing the hand.
This effect is most pronounced for functionally constrained objects (e.g., scissors, hammers), where task-semantic encoding ensures the handle faces the receiver.

FC-Handover and Handover-VA receive comparable comfort and safety scores. FC-Handover yields mean scores of 3.2 for comfort and 2.9 for safety, whereas Handover-VA averages 2.8 in both metrics. However, they differ substantially in naturalness (FC: median\,=\,3.5, mean\,=\,3.4; VA: median\,=\,2, Q1\,=\,1, mean\,=\,2.5).
The fixed orientation heuristic in Handover-VA frequently presents objects at angles that require wrist rotation to complete the grasp, which we identify as a key factor limiting naturalness.
FC-Handover avoids this by predicting a full hand mesh proxy. This approach yields more varied orientations and higher naturalness, despite the method ranking third in simulation success rate (59.6\%).

Contact-Handover scores lowest in both comfort (median\,=\,2.5, mean\,=\,2.5) and perceived safety (median\,=\,2, Q3\,=\,3, mean\,=\,2.4).
When the predicted contact map misaligns under sensor noise, the method selects grasps that partially occlude the receiver's intended grasp region, forcing awkward repositioning or causing the object to slip; we note that such repeated failures significantly reduce perceived safety.
Even in successful trials, contact-optimized grasps occasionally orient the functional region away from the receiver.
Interestingly, naturalness (mean\,=\,2.9) was rated higher than comfort and safety, confirming that grasp pose occlusion rather than object orientation is the primary driver of low ratings.

These results partially refute the stated hypothesis.
Intent-Handover ranks first in both simulation SR (67.9\%) and user ratings across all three dimensions, consistent with the hypothesis.
However, Contact-Handover ranks second in simulation SR (66.2\%) yet receives the lowest comfort and safety ratings. Notably, no participant rated its safety above 4. Conversely, FC-Handover ranks third in simulation (59.6\%) but second in user preference.
This rank inversion demonstrates that simulation success rate alone is not a sufficient predictor of perceived handover quality.
Crucially, we observe a divergence between geometric and perceived safety: Contact-Handover achieved the lowest simulation safety failure rate ($F_{\text{safe}}=3.6\%$), yet users rated it least safe due to intrusive grasping behaviors.
In contrast, Intent-Handover's low S1 affordance failure rate ($F_{\text{afford}}=3.8\%$) strongly correlates with its superior naturalness ratings.
This confirms that while standard metrics like SR and $F_{\text{safe}}$ capture physical feasibility, higher-level metrics like $F_{\text{afford}}$ are better indicators of human-centric handover quality.
This validates the effectiveness of R2HandoverSim. By decomposing performance into granular metrics, it successfully identifies the specific factors that drive user preference, offering a more predictive evaluation than success rates alone.

\section{Conclusion}\label{sec:conclusion}

We presented R2HandoverSim, a standardized simulation benchmark for robot-to-human handovers.
Our evaluation of four baselines demonstrates the trade-offs between geometric stability and human-centric metrics.
Real-world experiments with 30 participants confirm that the method achieving the highest simulation success rate (Intent-Handover) also leads in real-world success, and that the benchmark reveals distinct sim-to-real transfer characteristics across methods.
Furthermore, our analysis of multiple metrics shows that detailed metrics such as $F_{\text{afford}}$ predict perceived handover quality better than success rate alone.
Future work will incorporate dynamic hand models and closed-loop receiver policies to further bridge the sim-to-real gap.

\bibliographystyle{IEEEtran}
\bibliography{reference}

\end{document}